\documentclass{article} 
\usepackage[final]{nips_2017}
\usepackage{natbib}
\usepackage{times}
\usepackage{hyperref}
\usepackage{url}
\usepackage{graphicx}
\usepackage{amsmath}
\usepackage{algorithm, algpseudocode}
\usepackage{grffile} 
\usepackage{afterpage}
\usepackage{booktabs}
\usepackage{color}
\usepackage{verbatim}
\usepackage{soul}
\usepackage{xspace}
\usepackage{comment}
\usepackage{array}
\usepackage{multirow}
\usepackage{url}
\usepackage{epigraph}

\usepackage{caption}
\usepackage{subcaption}
\usepackage{graphicx}

\usepackage{amsmath}
\DeclareMathOperator{\sgn}{sgn}

\begin{document}
\title{Be Careful What You Backpropagate: A Case For Linear Output Activations \& Gradient Boosting}

\author{
Anders {\O}land
\quad
Aayush Bansal
\quad
Roger B. Dannenberg
\quad
 Bhiksha Raj \\\\
School of Computer Science\\
 Carnegie Mellon University\\
 Pittsburgh, PA 15213 \\
\texttt{\{anderso,aayushb,rbd,bhiksha\}@cs.cmu.edu}
}

\maketitle

\begin{abstract}
In this work, we show that saturating output activation functions, such as the softmax, impede learning on a number of standard classification tasks. Moreover, we present results showing that the utility of softmax does not stem from the normalization, as some have speculated. In fact, the normalization makes things worse. Rather, the advantage is in the exponentiation of error gradients. This exponential gradient boosting is shown to speed up convergence and improve generalization. To this end, we demonstrate faster convergence and better performance  on diverse classification tasks: image classification using CIFAR-10 and ImageNet, and semantic segmentation using PASCAL VOC 2012. In the latter case, using the state-of-the-art neural network architecture, the model converged \textbf{33\%} faster with our method than with the standard softmax activation, and that with a slightly better performance to boot.

\end{abstract}

\section{Introduction}
Training a deep neural network (NN) is a highly non-convex optimization problem that we usually solve using convex methods. For each extra layer we
add to the network, the problem becomes more non-convex, i.e. more curvature is added to the error surface, making the optimization harder.
Yet, it is commonplace to add unnecessary curvature at the output layer even though this does not expand the space of functions that the NN can represent. This curvature is then back-propagated through all the previous layers, causing a detrimental increase in the number of ripples in the error surfaces of especially the lower layers, which are already the toughest ones to train. This is done, in part, so that we can all pretend that the outputs are probabilities, even though they really are not. 
In the following, we show that saturating output activation functions, such as the softmax, impede learning on a number of standard classification tasks. Moreover, we present results showing that the utility of the softmax does not stem from the normalization, as some have speculated [\cite{Goodfellow2016DeepLearning, Keck2012FeedforwardCoin}]. In fact, the normalization makes things worse. Rather, the advantage is in the exponentiation of error gradients. This exponential gradient boosting is shown to speed up convergence and improve generalization.

\subsection{Squashers \& Saturation}
Historically, output squashers, such as the logistic (aka sigmoid) and tanh functions, have been used as a simple way of reducing the impact of outliers on the learned model. For example, if you fit a model to a small dataset with a good amount of outliers, those outliers can produce very large error gradients that will push the model towards a hypothesis that favors said outliers, leading to poor generalization. Squashing the output will reduce those large error gradients, and thus reduce the negative influence of the outliers on the learned model. However, if you have a small dataset, you should not use a neural network in the first place---other methods are likely to work better. And if you have a large dataset, the impact of any outliers will be minuscule. Therefore, the outlier argument is not very relevant in the context of deep learning. What is relevant, however, is that squashing functions saturate, resulting in very small gradients, appearing in the error surface as infinite flat plateaus, that slow down learning, and even cause the optimizer to get stuck [\cite{LeCun2012, Glorot2010}]. This observation was part of the motivation behind applying the now popular ReLU activation (rectified linear units) to convolutional neural nets [\cite{Krizhevsky2012, NairRectifiedMachines, Jarrett2009WhatRecognition}]. Surely, the massive success of ReLUs (and other related variants) speaks to the importance of avoiding saturating activations. Yet, somehow this knowledge is currently not being applied at the output layer! We contend, that for the very same reason that saturating units in the hidden layers should be avoided, linear output activations are to be preferred. 

\subsection{The Softmax Function}
The most famous culprit, among the saturating non-linearities, is of course the softmax function [\cite{BridleProbabilisticRecognition}], $y_j = \frac{\exp(x_j)}{\sum_i \exp(x_i)}$. This is the de facto standard activation used for multi-class classification with one-hot target vectors. Even though it is technically not a squasher, but a normalized exponential, it suffers from the same problem of saturation. That is, when the normalization term (the denominator) grows large, the output goes towards zero.          
The original motivation behind the softmax function was not dealing with outliers, but rather to treat the outputs of a NN as probabilities conditioned on the inputs. As intriguing as this may sound, we must remember that in most cases the outputs of the softmax would actually \emph{not} be true probabilities. To claim that outputs are probabilities, we must assume a within-class Gaussian distribution of the data, made in the derivation of the function [\cite{BridleProbabilisticRecognition}]. In practice, we say that the outputs may be interpreted as probabilities, as they lie in the range $[0;1]$ and sum to unity [\cite{Bishop1995NeuralRecognition, Bishop2007PatternEdn}]. However, if these are sufficient criteria for calling outputs probabilities, then the normalization might just as well be applied after training, which would not make the probabilistic interpretation any less correct. This way, we can avoid the problem of saturation during training, while still pretending that the outputs are probabilities (in case that is relevant to the given application). Another potential drawback of the normalization is that it bounds the function at both ends s.t. $f:\mathcal{R}\to [0;1]$. Consequently, when we apply it at the output layer, s.t. $y=f(x)$, where the error gradient (or ``error delta'') ${\frac{\nabla \mathcal{L}(y,t)}{\nabla x}} = y-t$, and $t\in\{0,1\}$, we effectively bound the gradients too, which affects all the previous layers during back-propagation.   

\subsection{The Main Idea}
Simply put, our main idea is to apply a bit of common sense to the aforementioned situation. Namely, that we are solving highly non-convex optimization problems using a convex method: backpropagation [\cite{Rumelhart1986LearningErrors}] with stochastic gradient descend (SGD). Even in saying those words, it appears evident that making the problem more non-convex---for no good reason---has to be a bad idea. Following that simple logic, output activations should \emph{always} be linear (the identity function) unless there is some advantage in adding non-linearity that somehow outweighs the price that must be paid in added non-convexity. Thus, we take the view that the only things that really matter are the speed of convergence and the final classification accuracy. We do not care about probabilistic interpretations, loss functions, or even, to some extent, mathematical correctness. Training a neural network is the process of iteratively pushing some weights in the right direction, and while doing so, we want to make the most of what we have: our error gradients. This does not entail imposing pointless bounds on them, or allowing them to become very small for no good reason. 

Table \ref{tab:mnist} shows what happened when we first applied this view on real data; the MNIST dataset [\cite{LeCun1998GradientRecognition}]. Training a simple three-layer NN (fully connected) with ReLUs in the hidden layers, we compared the median results obtained over twenty trials with sigmoid, tanh, and linear output activations. The learning rate was fixed, and carefully tuned for each setting, and neither dropout [\cite{Hinton2012ImprovingDetectors}], batch normalization [\cite{Ioffe2015}], nor weight decay was used. The NN trained for 100 epochs, and the point of convergence is set to be the epoch where the minimum classification error was observed. This experiment was repeated multiple times with other hidden activations, and weight initialization schemes, and they all gave the same result: with linear output activations, the rate of convergence is reduced by approximately 25 percent (and some moderate improvements in generalization was observed as well). Note, that the softmax is not included in the table for the very simple reason that it gave miserable results on this NN configuration. 

\begin{table}
\setlength{\tabcolsep}{3pt}
\def\arraystretch{1.2}
\center
\begin{tabular}{@{}l c c c }
\toprule
  \textbf{Output Activation} & \textbf{Error} & \textbf{Convergence}  \\
\midrule
Sigmoid   & 1.8 &   98.5       	\\
Tanh      & 1.7 &   95.0      	\\
Linear    & 1.7 &   \textbf{73.5}         \\
\bottomrule
\end{tabular}
\caption{Median results (20 trials) on MNIST for a 392-50-10 NN with ReLUs in the hidden layers; final classification error \& no. of epochs needed to converge.}
\label{tab:mnist}
\end{table}

\section{Gradient Boosting}
When we first tried to train a convolutional neural network (CNN) on the CIFAR-10 data [\cite{Krizhevsky2009LearningImages}], with linear instead of softmax outputs, we expected to see at least a hint of improvement. This was not the case. On the contrary, the softmax clearly won that battle. The reason for this lies in the exponentiation of the outputs. For a moment, stop thinking about the softmax in a probabilistic context, and instead view it as the equivalent of linear outputs, with a mean squared error loss, combined with non-linear boosting of the error deltas, $y - t$. From this perspective, it becomes clear that when we have $y_j = \frac{\exp(x_j)}{\sum_i \exp(x_i)}$, nothing changes with respect to the one-hot classification, but large errors will be exponentiated. This allows the optimizer to take bigger steps towards a minimum, thus leading to faster convergence. An intuitive interpretation of this would be that when we are confident about an error, we can take an exponentially larger step towards minimizing that error. The idea bears some resemblance to momentum, where we gradually speed things up when the error gradients are consistent.

\subsection{Exponential Boosting}
If exponentiation of error deltas is good, and saturation is bad, it follows that using an ``un-normalized'' softmax, so to speak, should yield an improvement. That is, simply use linear outputs, $y = x$, but compute the error gradients as ${\frac{\nabla \mathcal{L}(y,t)}{\nabla x}} = \alpha \exp(y) - t$. Alternatively, we can think of it as an exponential output activation with an incorrect gradient formulation imposed on it, i.e. $y - t$ (this is in fact how we implemented it). As seen in Figure \ref{fig:cnn5_median}, this simple change does in fact lead to a consistent boost in performance. The result was obtained on CIFAR-10 with a 5-layer CNN; four convolutional layers followed by an affine output layer with linear outputs and exponential gradient boosting (exp-GB), and batch normalization in all layers. We set $\alpha = 0.1$, which has worked well in all our experiments; sometimes $\alpha = 0.01$ is slightly better. To further boost the non-linear interaction between the outputs and the targets, we used larger target values, $t \in \{0,16\}$ instead of $t \in \{0,1\}$. As can be seen in the histograms of Figure \ref{fig:hist} (from a different experiment), this produces much larger gradients. The deltas are roughly in $[-6;10]$, as opposed to the bounded errors of the softmax, that are in $[-1;1]$. The idea of picking better target values is not new. To reduce the risk of saturating with logistic units, \cite{LeCun2012} recommend choosing targets at the point of the maximum of the second derivative.

Another potential advantage of the exponentiation is that $exp(x)$ asymptotically approaches zero towards negative infinity. This is
especially advantageous with one-hot target vectors, because we do not care about exact output values as long as the correct class has the largest value. Hence, we can mostly ignore any negative errors in outputs for the negative classes. This can be seen as a relaxation of the optimization problem, where we are essentially trying to solve an inequality for the negative classes instead of an exact equality. With linear activations \emph{without} exp-GB, the optimizer would always try to push the outputs for the negative classes towards zero. This can lead to situations where an otherwise correct output (i.e. the maximum value belongs to the node representing the target class) for a given input, $x_i$, leads to a weight update that renders the output incorrect the next time that $x_i$ is seen; this is in exchange for the mean output for the negative classes being slightly closer to zero than on the previous iteration. That is a bad result, but we avoid this problem when we exponentiate our gradients.       

\subsection{Cubic Boosting}
\begin{figure*}[t!]
\begin{subfigure}[t]{0.5\textwidth}
    \centering
    \includegraphics[scale=.5]{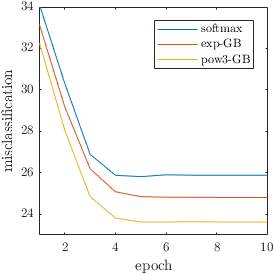}
    \caption{Median misclassification rates (20 trials) for a 5-layer CNN with softmax outputs, linear outputs with exponential boosting (exp-GB), and linear outputs with cubic boosting (pow3-GB).}
    \label{fig:cnn5_median}
\end{subfigure}
~
\begin{subfigure}[t]{0.5\textwidth}
    \centering
    \includegraphics[scale=.5]{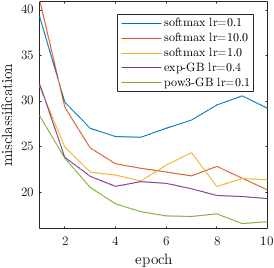}
    \caption{Misclassification rates for a 10-layer CNN with softmax outputs (for three different learning rates), exponential boosting (exp-GB), and cubic boosting (pow3-GB).}
    \label{fig:cnn10_mis}
\end{subfigure}
\caption{Classification Errors on CIFAR-10}
\end{figure*}

Although we can often ignore large negative outputs that yield large negative error deltas, we cannot ignore all of them. This raises the question whether we may further boost performance by also allowing for the exponentiation of large \emph{negative errors}. The answer is: yes we can! An obvious candidate would be a mirrored exponential function, $y = \alpha \sgn(x) exp(\lvert x \rvert - 1) + \beta$, where $\sgn(\cdot)$ is the sign function. However, this function does not have a nice and cozy place for us to put all those gradients that we do not need to worry about, so it does not work well. Instead, we use a simple polynomial that has a conveniently flat area around $[-1; 1]$, $y= \alpha x^3 + \beta$; let's call it pow3-GB. Taking another look at Figure \ref{fig:cnn5_median}, we see that this does
indeed work better; following exactly the same trend as observed with exp-GB, that the error drops significantly faster than with the softmax. We set $\alpha = 0.001$, $\beta = 0.4$, and use target values $t \in \{0, 10\}$.

\label{fig:mis}

\section{Experiments}

We now carefully study the performance of gradient boosting (GB) for image-classification on CIFAR-10 [\cite{Krizhevsky2009LearningImages}] and ImageNet 2012 [\cite{Russakovsky15}], and the pixel-level task of semantic segmentation on the PASCAL VOC 2012 dataset [\cite{Everingham10}].

\textbf{CIFAR-10 Classification.} In this experiment, our purpose is not to get state-of-the-art results but rather to learn how increased depth may affect our method. We use an (almost) all convolutional network with ten layers; following the principle presented in [\cite{SpringenbergDBR14}], but with batch normalization, and the average pooling layer replaced by a fully-connected one. The latter was done to make computation more deterministic, so as to allow for better evaluation of the effects of changing various parameters. Note that pooling involves atomic operations on the GPU, which can result in relatively large variance in output. For this experiment, we used a fixed learning rate and carefully tuned it with the purpose of getting the best result within ten epochs.  We use the same $\alpha$ and $\beta$ values as in our previous experiment, but this time we use different target values. $t \in \{0,6\}$ produced better results for exp-GB. With pow3-GB it seemed a good idea to try negative target values for the negative classes since the function is not bounded at the lower end; we saw a significant improvement when using $t \in \{-2,10\}$.

\begin{figure*}[t!]
    \centering
    \begin{subfigure}[t]{0.5\textwidth}
        \centering
        \includegraphics[scale=0.65]{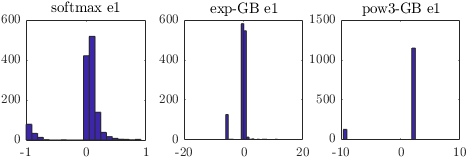}
    \end{subfigure}%
    \\
    \begin{subfigure}[t]{0.5\textwidth}
        \centering
        \includegraphics[scale=0.65]{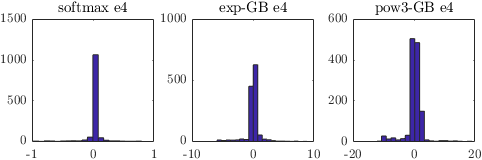}
    \end{subfigure}%
    
    \caption{{\bf Top:} The distribution of output layer error gradients for softmax, linear with exponential boosting, and linear with cubic boosting at the start of epoch 1; training a 10-layer CNN on CIFAR-10. {\bf Bottom}: same, but for epoch 4.}
    \label{fig:hist}
\end{figure*}

\begin{figure}
    \centering
    \includegraphics[scale=.58]{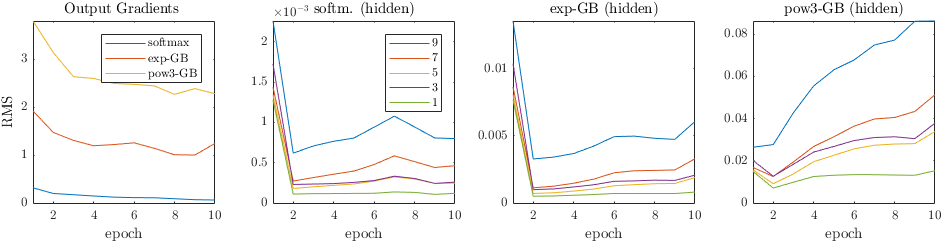}
    \caption{RMS of error gradients over ten epochs of training a 10-layer CNN on CIFAR-10. {\bf Left:} output layer. {\bf Rest:} every second hidden layer. Note the upwards trend from epoch 2 onwards in the hidden layers for GB.}
    \label{fig:rms_hid}
\end{figure}

Figure \ref{fig:cnn10_mis} shows how the classification error evolved during training. For softmax, we show results from trying three different learning rates to insure that our choice of $1.0$ really is a good one. We note that the overall trend is the same as for the 5-layer CNN; for the first 2-5 epochs, the error rates drop significantly faster with GB than with the softmax. The histograms in Figure \ref{fig:hist} show the distribution of the output error deltas for the first batch of epoch 1 and epoch 4. The larger target values used for GB are clearly reflected; resulting in sharper distributions clustered around the negated target values. This is of course most significant on the first iteration, but the trend is still very clear in the fourth (and tenth) epoch. This boosting of the output errors has a very significant effect on the gradient signals received in the hidden layers during backpropagation. Figure \ref{fig:rms_hid} shows this effect very nicely via the root mean square (RMS) of the gradients. With exp-GB, the RMS of the hidden layer gradients is an order of magnitude higher than with the softmax; for pow3-GB it is more than two orders of magnitude. Interestingly, the hidden-layer RMS-gradients recorded for pow3-GB grow rapidly from the second epoch and onwards. A similar trend is seen for exp-GB, albeit less dramatically, and for the softmax there is only a slight upwards trend, and only in the top layers. This correlates well with the error rates (see Figure \ref{fig:cnn10_mis}); the softmax gets stuck early on, and the linear activations with gradient boosting continue to learn through all ten epochs. All in all, this seems to indicate that gradient boosting may help alleviate the infamous problem of vanishing gradients [\cite{Hochreiter1991UntersuchungenNetzen,Goodfellow2016DeepLearning}] in deep neural networks.

\textbf{ImageNet Classification.} The ImageNet 2012 dataset [\cite{Russakovsky15}] consists of \~1.3 million RGB images that are much larger than the tiny images of CIFAR-10. Training a state-of-the-art model on this data can take weeks. Thus, for this experiment, we will again consider only the first ten epochs of training on a relatively shallow and well-known architecture, AlexNet [\cite{Krizhevsky2012}]. Figure \ref{fig:imagenet} and Table \ref{tab:imagenet} show the median classification errors over five trials. With exp-GB, we get a median reduction in the top5-error of 4.52 percent, and a 3.74 percent reduction of the top1-error; i.e. the minimum errors achieved within ten epochs. Thus, the result follows the general trend of our previous experiments. However, there are two important differences. First, the pow3-GB did not work well, whereas exp-GB clearly still outperformed the softmax. Secondly, we had to use batch normalization (BN) to get good results.

With respect to the failure of pow3-GB, the explanation is likely found in the 100-fold increase in the number of classes; compared to the ten classes of CIFAR-10. Because such large error gradients are back-propagated from the output layer, the errors in the hidden layers simply blow up too much, when one thousand errors are multiplied and summed instead of just ten. In a way, this is the opposite problem of what we could expect to see with the softmax, where the saturation is likely to be worse with more classes (as the normalization term grows), thus producing very small gradients. With respect to why BN becomes more important, again, the reason is that the magnitude of the back-propagated gradients depends on the number of classes. The larger gradients result in bigger and faster changes in the distributions of the activations in the hidden layers; thus magnifying internal covariate shifts, and increasing the need for BN. 

For this experiment, we used (carefully tuned) fixed learning rates of 0.01 and 0.001 for the softmax and exp-GB, respectively. For exp-GB we set $\alpha = 0.01$ and used target values, $t \in \{0,10\}$.

\begin{figure*}[t!]
\begin{subfigure}[t]{0.5\textwidth}
    \centering
    \includegraphics[scale=.5]{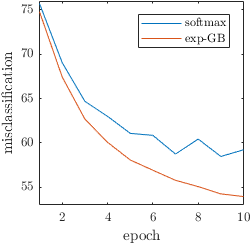}
    \caption{{\bf Top1-Error:} AlexNet with BN.}
    \label{fig:top1}
\end{subfigure}
~
\begin{subfigure}[t]{0.5\textwidth}
    \centering
    \includegraphics[scale=.5]{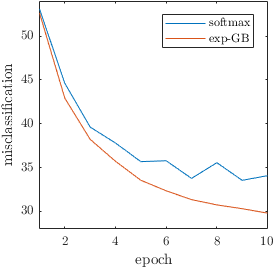}
    \caption{{\bf Top5-Error:} AlexNet with BN.}
    \label{fig:top5}
\end{subfigure}
\caption{Median Classification Errors on ImageNet 2012}
\end{figure*}
\label{fig:imagenet}

\begin{table}
\setlength{\tabcolsep}{3pt}
\def\arraystretch{1.2}
\center
\begin{tabular}{@{}l c c c }
\toprule
  \textbf{Output Activation} & \textbf{Top1-Error} & \textbf{Top5-Error}  \\
\midrule
Softmax      & 58.47 &   33.55      	\\
Exp-GB    & \textbf{53.95} &   \textbf{29.81}         \\
\bottomrule
\end{tabular}
\caption{Minimum error reached (median over 5 trials) training AlexNet with BN on ImageNet 2012.}
\label{tab:imagenet}
\end{table}

\textbf{Semantic Segmentation.} We now evaluate our method for the pixel-level classification task of semantic segmentation. The goal in semantic segmentation is to determine class labels for every single pixel in an image. Prior work [\cite{PixelNet,Hariharan15,Long15}] in this direction use a fully convolutional network with the standard softmax and multi-class cross-entropy loss for optmization. In this experimwent, we use the PixelNet architecture [\cite{PixelNet}]. This model uses a VGG-16 [\cite{SimonyanZ14a}] architecture (pre-trained on ImageNet) followed by a multi-layer perceptron that is used to do per-pixel inference over multi-scale descriptors. It has achieved state-of-the-art performance for various pixel-level tasks such as semantic segmentation, depth/surface normal estimation, and boundary detection. 

We evaluate our findings on the heavily benchmarked Pascal VOC 2012 dataset. Similar to prior work [\cite{PixelNet,Hariharan15,Long15}], we make use of additional labels collected on 8498 images by \cite{hariharan11}. We keep a small set of 100 images for validation to analyze convergence, and use the same settings as used for analysis in [\cite{PixelNet}]: a single scale $224{\times}224$ image is used as input to train the model. All the hyper-parameters are kept constant except the initial learning rate\footnote{The initial $lr=1{\times}10^{-3}$ for softmax, $lr=1{\times}10^{-4}$ for exp-GB, and $lr=5{\times}10^{-5}$ for pow3-GB. Lowering the learning rate for softmax deteriorates the performance.}. We report results on the Pascal VOC-2012 test set (evaluated on the PASCAL server) using the standard metrics of region intersection over union (\textbf{IoU}) averaged over classes (higher is better). 
 
 Table~\ref{tab:voc_2012} shows our results (both per-class and \textbf{mIoU}) for GB and the standard softmax. We observe that the model trained using {\em exp-GB} converged after 40 epochs, whereas the {\em softmax} model converged after 60 epochs. As seen in Table~\ref{tab:voc_2012}, our method provides \textbf{33\%} faster convergence, while yielding a slightly better performance (\textbf{E-40} vs. \textbf{S-60}). We see a significant \textbf{3\%} boost in the first 40 epochs with exp-GB (\textbf{E-40} vs. \textbf{S-40}). 
 
 Additionally, recent work [\cite{varol17,Wang15}] in the computer vision community have formulated regression problems such as depth and surface normals estimation, in a classification paradigm, in hope of easier optimization and better performance. From these experiments, we however infer that it is likely not the {\em softmax$+$cross-entropy} that boosts the performance. Rather, it is the use of one-hot encoding of the target vectors.
  
 \begin{table*}[t!]
\tiny{
\setlength{\tabcolsep}{3pt}
\def\arraystretch{1.2}
\center
\begin{tabular}{@{}l c c c c c c c c c c c c c c c c c c c c c@{}p{0.3cm}@{}c@{}}
\toprule
  & bg & aero  &   bike &  bird & boat &  bottle  &  bus  &  car  &  cat  &  chair & cow & table &  dog  & horse & mbike & person  & plant & sheep & sofa & train & tv &  &\textbf{mIoU}\\
\midrule
\textbf{E-40}  & 92.3 &
\textbf{87.9} &       \textbf{43.2} &        73.6 &        54.6 &       \textbf{68.8} &        83.9 &        \textbf{82.2} &        77.1 &        \textbf{27.7} &        62.0 &       50.8 &        \textbf{74.6} &        \textbf{75.5} &        \textbf{80.6} &        \textbf{78.7} &        47.4 &       \textbf{74.0} &        43.2 &       \textbf{76.0} &        \textbf{60.1} &       &  \textbf{67.3} \\
\textbf{P-40}  & 92.3 &
 86.0 & 39.1        & \textbf{74.1}        &     49.4   &    66.3    & \textbf{84.7}        &     79.7   &    77.4     & 26.4       &     63.2   &    51.6    &   71.4     &      74.7  &     79.8   &    76.6     &  45.1      &     70.6   &    47.5    &   71.8      &     59.9    &     & 66.1  \\
\textbf{S-40}  & 91.9 &
84.9 &        38.5 &        66.8 &        54.0 &        63.4 &        79.8 &        72.9 &        72.7 &        25.4 &        63.6 &     55.4 &        68.2 &        72.7 &        75.5 &        76.2 &        46.7 &       71.6 &        42.8 &        71.2 &        58.4 &       &  64.4 \\
\midrule
\textbf{S-60}  & \textbf{92.4} & 
86.7 &        39.8 &        72.4 &        \textbf{58.0} &        65.6 &        82.9 &        78.9 &       \textbf{77.8} &        26.6 &       \textbf{66.1} &        \textbf{59.2} &       71.6 &        74.2 &       77.5 &        77.1 &        \textbf{49.3} &        73.8 &       \textbf{45.7} &       73.9 &        58.4 &       &  67.1 \\
\bottomrule
\end{tabular}
\vspace{0.2cm}
\caption{\textbf{Evaluation on Pascal VOC-2012 for Semantic Segmentation:} We found our analysis consistent for the pixel-level task of semantic segmentation. With only 40 epochs, our formulation exceeds the performance using Softmax$+$Cross-Entropy for 60 epochs. \textbf{E} deonotes exp-GB$+$mean-squared-error; \textbf{P} denotes pow3-GB$+$mean-squared-error; and \textbf{S} denotes softmax$+$cross-entropy-loss.}
\label{tab:voc_2012}
}
\vspace{-0.1cm}
\end{table*}

\textbf{Summary.} We evaluated our findings on two standard tasks of classification, i.e. image classification and pixel-level classification, on heavily benchmarked datasets. We observe a consistent trend for all these tasks: (1). softmax impedes learning; (2). exp-GB$+$mean-squared-error converges \textbf{25-35\%} faster than standard softmax$+$cross-entropy, and that too with a slightly better performance (not at the cost of it). We believe our results are important not just from a convergence perspective, but also from the point-of-view of having a general loss function for both classification and regression tasks.

\section{Further Analysis}
We can take a slightly more theoretical view on gradient boosting, and why it speeds up the convergence, by reasoning about second-order properties of the error surfaces induced by exp-GB and pow3-GB. This is typically done with the Hessian matrix, $H$, of second derivatives, which tells us something about the rate of change in the error for a single step of gradient descent. To keep things simple, we will consider only the case of a single output activation, i.e. a single dimension, so we do not need the full Hessian, $\frac{d^2}{dx^2}f$ will do. We will look at $\frac{\partial^2 E}{\partial x^2}$, where $E$ is the sum-of-squares error, $E(y, t) = \frac{1}{2} \sum_i \lVert y - t \rVert^2$. For our purpose we can simply ignore the summation in $E$, and just analyze $\frac{\partial^2 E}{\partial x^2}$ for a single example, $(x,t)$. Let us start by comparing the Hessians for linear, softmax, exponential, and cubic activations.       

For a linear activation, $y=x$, the Hessian is simply
\begin{align}
    H_{linear} = {\frac{\partial^2 \lVert y - t \rVert^2}{\partial x^2}} = 1
\end{align}

Re-writing the softmax activation as $y = \frac{e^x}{s}$, where $s = \sum_i e^{x_i}$ is a proxy for the normalization term, we get

\begin{align}
    H_{softmax} &= {\frac{\partial^2 \lVert {\frac{e^{x}}{s}} - t \rVert^2}{\partial x^2}}
    &= {\frac{e^{2x}}{s^2}} - \frac{e^x}{s} (t - \frac{e^x}{s})
\end{align}

and for exponential and cubic activations we have,

\begin{align}  
    H_{exp} &= {\frac{\partial^2 \lVert e^{x} - t\rVert^2}{\partial x^2}} &=
    e^{2x} - e^x (t - e^x)    
    \\
    H_{pow3} &= {\frac{\partial^2 \lVert x^3 - t\rVert^2}{\partial x^2}} &=
    9x^4 - 6x(- x^3 + t)
\end{align}

If we consider the situation where $x$ is near some local minimum, we know that the error surface will be locally convex around that point. This means that $H \approx 0$, and that the first and second term in each of the above Hessians will be approximately equal (i.e. they cancel each other out). Thus, we will ignore the second terms, and simply compare the growth of all the first terms, as we move $x$ away from that local optimum. Now it becomes immediately evident that (locally) $H_{pow3} > H_{exp} > H_{softmax} > H_{linear}$, because as $x \to \pm \infty$ we get $9x^4 > e^{2x} > {\frac{e^{2x}}{s^2}} > 1$ for all $s > 1$. Unsurprisingly, it all depends on the magnitude of the normalization term of the softmax, $s = \sum_i e^{x_i}$. If $s$ is very small $H_{softmax}$ will blow up, so we need to assert the probability of that happening. At the onset of training, it is reasonable to assume that the input to the softmax will be evenly distributed around zero. Thus, half of the $x_i$'s are positive, guaranteeing that $s>1$ as $\forall x>0, e^x > 1$. To see what happens later, we can consider a numerical example for one thousand classes. Even when the model is trained well, such that the $x_i$'s for the 999 negative classes are likely to be negative and contribute very little to $s$ as $e^{x_i} \ll 1$---it still takes only one single $x_i \ge 0$ to make $s>1$ (likely to be the one for the positive class). It seems reasonable to claim that this will probably be the case most of the time.

\begin{figure}
    \centering
    \includegraphics[scale=.5]{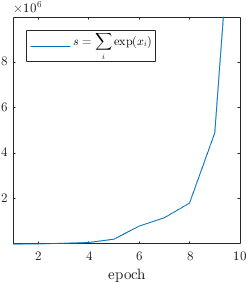}
    \caption{Magnitude of the softmax normalization term, $s = \sum_i e^{x_i}$, recorded for a 10-layer CNN trained on CIFAR-10. It starts out with $s = 2,342$ and blows up from there.}
    \label{fig:sum}
\end{figure}

To back up this claim, we take a look at the actual $x_i$'s recorded during training of the 10-layer CNN from our CIFAR-10 experiment in the previous section. Figure \ref{fig:sum} shows how the normalization term, $s$, of the softmax actually behaved. It starts out with a value of 2,342 and increases monotonically from there. 

However, we need to remember that for GB the Hessians are a little different, as we are just boosting the error gradients, $y-t$. Thus, the second derivatives are just the derivatives of those deltas, with $H_{exp-GB}= e^x$, $H_{pow3-GB}=3x^2$, and $H_{softmax}=1$ (with the multi-class cross-entropy loss)---which only adds to our point that GB can minimize the error faster than the softmax.

\section{Conclusion}
Our results suggest fundamental changes in deep network training, and to our perception of the omnipresent softmax function. In a way, all that we have done is to apply common sense to the challenge of training deep non-convex models using the convex method of gradient descent. Specifically, do not make the problem more non-convex that it needs to be. Whenever we add curvature to the error surface, we make the optimization harder, and we should always keep this in mind when we make decisions on how we configure our models during training. 

Taking the consequence of this, by e.g. skipping the normalization term of the softmax, we get significant improvement in our NN training---and at no other cost than a few minutes of coding. The only drawback is the introduction of some new hyper-paramters, $\alpha$, $\beta$, and the target values. However, these have been easy to choose, and we do not expect that a lot of tedious fine-tuning is required in the general case. 

From this perspective, our work---and much of literature---is concerned with treating the symptoms rather than the cause. The cause of our problems is our use of gradient-based optimizers. Perhaps one day we will have a better learning algorithm, but until we do: be careful what you back-propagate!

\bibliographystyle{apalike}
\bibliography{gradboost_nips2017}

\end{document}